# The Cost of Perfect English: Pragmatic Flattening and the Erasure of Authorial Voice in L2 Writing Supported by GenAI

Ao Liu, Shanhua Zhu
School of Foreign Languages, Southeast University

**Abstract:** The rapid integration of Generative AI (GenAI) into language teaching and learning offers second language (L2) writers unprecedented tools for text optimization. However, the pursuit of native-like academic fluency often comes at the cost of sociopragmatic diversity. Investigating the phenomenon of "pragmatic flattening"—the systematic erasure of culturally preferred politeness strategies and authorial stance—this study conducts a large-scale comparative analysis of argumentative essays written by Chinese B2-level university students from the ICNALE corpus. The original texts were polished using the official APIs of four leading Large Language Models with a strict zero-temperature setting to ensure reproducibility. Preliminary findings reveal a nuanced "dimensional divergence" within the Semantic Preservation Paradox. While the models successfully corrected lexicogrammatical errors and retained core propositional meaning, their interventions in the sociopragmatic layer were bifurcated. In the interactive dimension, all models exhibited a drastic and universal collapse of dialogic engagement markers, transforming negotiated discourse into monologic assertions. Conversely, in the epistemic stance dimension, the models demonstrated architecture-based variability: some models aggressively scrubbed epistemic markers, while others reinforced tentative hedging as a form of decontextualized algorithmic caution. This phenomenon confirms the Semantic Preservation Paradox, demonstrating that GenAI, while enhancing accuracy, systematically overwrites the unique rhetorical identities of L2 writers into a homogenized Anglo-American academic paradigm. The study argues that future instructional interventions must move beyond mere error correction, advocating for the integration of Critical AI Literacy to empower multilingual writers to utilize GenAI for linguistic enhancement while safeguarding their sociopragmatic diversity and rhetorical agency.

# 1.Introduction

The rapid integration of Generative Artificial Intelligence (GenAI) into educational ecosystems has fundamentally transformed second language (L2) writing pedagogy. For multilingual learners, GenAI tools, particularly Large Language Models (LLMs), offer unprecedented affordances for text optimization, providing immediate, near-native language modeling. Recent empirical studies underscore the efficacy of these models in language learning contexts. For instance, Mizumoto et al. (2024) demonstrated GenAI's remarkable alignment with human raters in evaluating and improving L2 writing accuracy, while Strobl et al. (2025) highlighted how learners increasingly rely on these tools not merely as spell-checkers, but as collaborative agents in their revision and deliberation processes. At first glance, this technological shift seemingly solves the long-standing pedagogical challenge of achieving surface-level linguistic proficiency in foreign language practice.

However, the pursuit of near-native academic fluency often conceals a profound sociopragmatic cost. While GenAI excels at correcting lexicogrammatical errors, it simultaneously functions as a rigid normative filter. When L2 writers utilize AI to "polish" their texts, there is a severe risk of what this study conceptualizes as pragmatic flattening—the systematic erasure of culturally preferred politeness strategies, stance markers, and interpersonal engagement. Academic writing extends far beyond the mere transmission of propositional meaning; it necessitates the careful negotiation of authorial voice, epistemic stance, and genre-specific conventions (Zhao & Wu, 2024). Yet, as Kecskés and Dinh (2025) argue, current GenAI models often lack the intercultural pragmatic depth and nuanced negotiation strategies inherent in human meaning-making. By enforcing a singular, homogenized Anglo-American academic norm, GenAI inadvertently acts as an agent of linguistic standardization, leading to the "cultural ghosting" of the L2 writer's unique rhetorical identity.

This dynamic gives rise to a critical phenomenon identified in this paper as the Semantic Preservation Paradox: the AI successfully retains the core propositional meaning of the L2 text while aggressively scrubbing the interpersonal pragmatic markers that establish the writer's solidarity and negotiated stance. Despite a growing focus on the ethical use of language technologies, a significant blind spot remains in foreign language teaching regarding how this algorithmic flattening affects the face-saving strategies of learners, particularly those from high-context cultures. Current foreign language teaching practices often praise the grammatical accuracy achieved through AI, overlooking the algorithmic modification of face-threatening acts (FTAs) and epistemic hedging.

Addressing this critical gap, this study investigates the micropragmatic modifications made by GenAI when editing argumentative essays. Utilizing a controlled dataset of texts written by Chinese B2-level university students from the International Corpus Network of Asian Learners of English (ICNALE) and grounded in Brown and Levinson's Politeness Theory, this research meticulously tracks the algorithmic erasure of authorial voice. Ultimately, this paper argues that the widespread adoption of AI in education necessitates a paradigm shift. Moving beyond mere error-correction paradigms, the field must integrate Critical AI Literacy (Strobl et al., 2025) into L2 classrooms, empowering multilingual writers to utilize GenAI for linguistic enhancement without sacrificing their sociopragmatic diversity.

## 2.Literature Review

### 2.1 GenAI Integration in L2 Writing: The Focus on Lexicogrammatical Optimization

The integration of Generative Artificial Intelligence (GenAI) has fundamentally restructured second language (L2) writing pedagogy, shifting the paradigm from traditional automated evaluation to dynamic text generation and collaborative revision (Yang & Li, 2024). Recent empirical scholarship robustly validates GenAI's efficacy in enhancing linguistic accuracy. For instance, Mizumoto et al. (2024) and Kurt and Kurt (2024) demonstrated that tools like ChatGPT exhibit high reliability as automated feedback agents, effectively aligning with human raters in assessing and correcting L2 writing. Consequently, learners increasingly rely on these models not merely as spell-checkers but as collaborative agents for restructuring their evolving drafts and alleviating writing anxiety (Strobl et al., 2025; Li et al., 2025).

However, the prevailing research paradigm predominantly evaluates GenAI interventions through a functionalist lens, emphasizing surface-level linguistic correctness and structural coherence. As Wang et al. (2024) observed in their study on academic writing, the pursuit of near-native academic fluency often leads to texts that are structurally sound but highly standardized. This overemphasis on lexicogrammatical optimization inadvertently treats language as a culturally neutral medium. It creates a critical blind spot regarding how underlying algorithms, trained predominantly on Anglo-American rhetorical conventions, reshape the deeper sociopragmatic intentions embedded in student texts.

### 2.2 Pragmatics, Politeness, and the Erasure of Authorial Voice

To fully comprehend the hidden costs of this algorithmic standardization, L2 writing must be examined through the frameworks of cross-cultural pragmatics and Politeness Theory (Brown & Levinson, 1987). In academic discourse, L2 writers do not merely convey propositional facts; they actively construct contextual meanings and negotiate their rhetorical identities (Chen et al., 2024). According to Zhao and Wu (2024), L2 writers employ specific interactional metadiscourse—such as dialogic engagement strategies and tentative hedges—to mitigate Face-Threatening Acts (FTAs) and establish solidarity with their readers. These markers are conscious, culturally informed sociopragmatic choices rather than linguistic deficits.

Yet, recent evidence suggests that GenAI models inherently struggle to preserve these nuanced human markers. Barattieri di San Pietro et al. (2025) note the limitations of large language models in handling complex pragmatic inferences. When L2 learners utilize GenAI to "polish" their writing, the technology operates on optimization algorithms designed for maximum directness. Moon (2025) provided empirical evidence of this, showing how ChatGPT overrides the nuanced, negotiated stance of L2 students in email requests, replacing culturally informed politeness strategies with bald, highly direct formulations.

This algorithmic intervention leads to what Al Hosni (2025) describes as the loss of "linguistic fingerprints." GenAI systematically scrubs interpersonal markers, replacing the writer's original epistemic stance with a homogenized, authoritative academic voice (Zou et al., 2025). This study conceptualizes this pervasive phenomenon as pragmatic flattening—a process that enforces a singular pragmatic norm and inevitably results in the "cultural ghosting" of the multilingual learner's authentic voice.

### 2.3 The Pedagogical Shift: Toward Critical AI Literacy

The pervasive homogenization of L2 texts and the subsequent erasure of authorial voice raise profound ethical and pedagogical concerns. If language educators continue to praise the grammatical perfection achieved via AI without questioning its sociopragmatic impact, they risk delegitimizing the unique rhetorical identities of learners from diverse high-context cultures. Addressing this technological disruption requires moving beyond mere error-correction paradigms.

Recent scholarship advocates strongly for cultivating critical engagement with digital and AI technologies. Building upon foundational models of critical digital literacies in TESOL (Jiang & Gu, 2022; Darvin & Hafner, 2022), scholars now emphasize the urgent need for Critical AI Literacy (Darvin, 2025; Godwin-Jones, 2025). Multilingual writers must be explicitly empowered to recognize algorithmic biases, develop self-reflective mindsets, and consciously decide when to reject algorithmic modifications to preserve their sociopragmatic intent. However, to design effective pedagogical interventions, educators first need granular, empirical evidence of how precisely GenAI alters micropragmatic features in L2 academic writing.

### 2.4 Research Gap and Questions

While existing literature has broadly discussed the loss of human voice in AI-generated texts (Al Hosni, 2025; Zou et al., 2025), there is a conspicuous lack of micropragmatic comparative analyses tracking the exact algorithmic modification of Face-Threatening Acts (FTAs) and epistemic stance markers in L2 argumentative essays. Specifically, the "Semantic Preservation Paradox"—whereby AI successfully retains propositional meaning while aggressively scrubbing interpersonal pragmatic markers—remains under-investigated empirically.

To address this critical gap and provide an empirical foundation for Critical AI Literacy, this study poses the following research questions:

**RQ1:** How does GenAI systematically modify interpersonal pragmatic markers (specifically, dialogic engagement strategies and epistemic hedges) when polishing argumentative essays written by L2 English learners?

**RQ2:** To what extent does GenAI maintain the original sociopragmatic intent and face-saving strategies of L2 writers while correcting lexicogrammatical errors (i.e., the Semantic Preservation Paradox)?

**RQ3:** What are the pedagogical implications of "pragmatic flattening" for L2 writers' rhetorical identities, and how can these findings inform the development of Critical AI Literacy in language classrooms?

## 3. Methodology

To systematically investigate the phenomena of "pragmatic flattening" and the "Semantic Preservation Paradox" across different algorithmic architectures, this study employs a mixed-methods comparative design. The methodology integrates large-scale computational data extraction with highly controlled generative AI interventions and micropragmatic qualitative coding. The overall research workflow and experimental stages are visualized in Figure 1.

**Figure 1. Experimental design workflow**

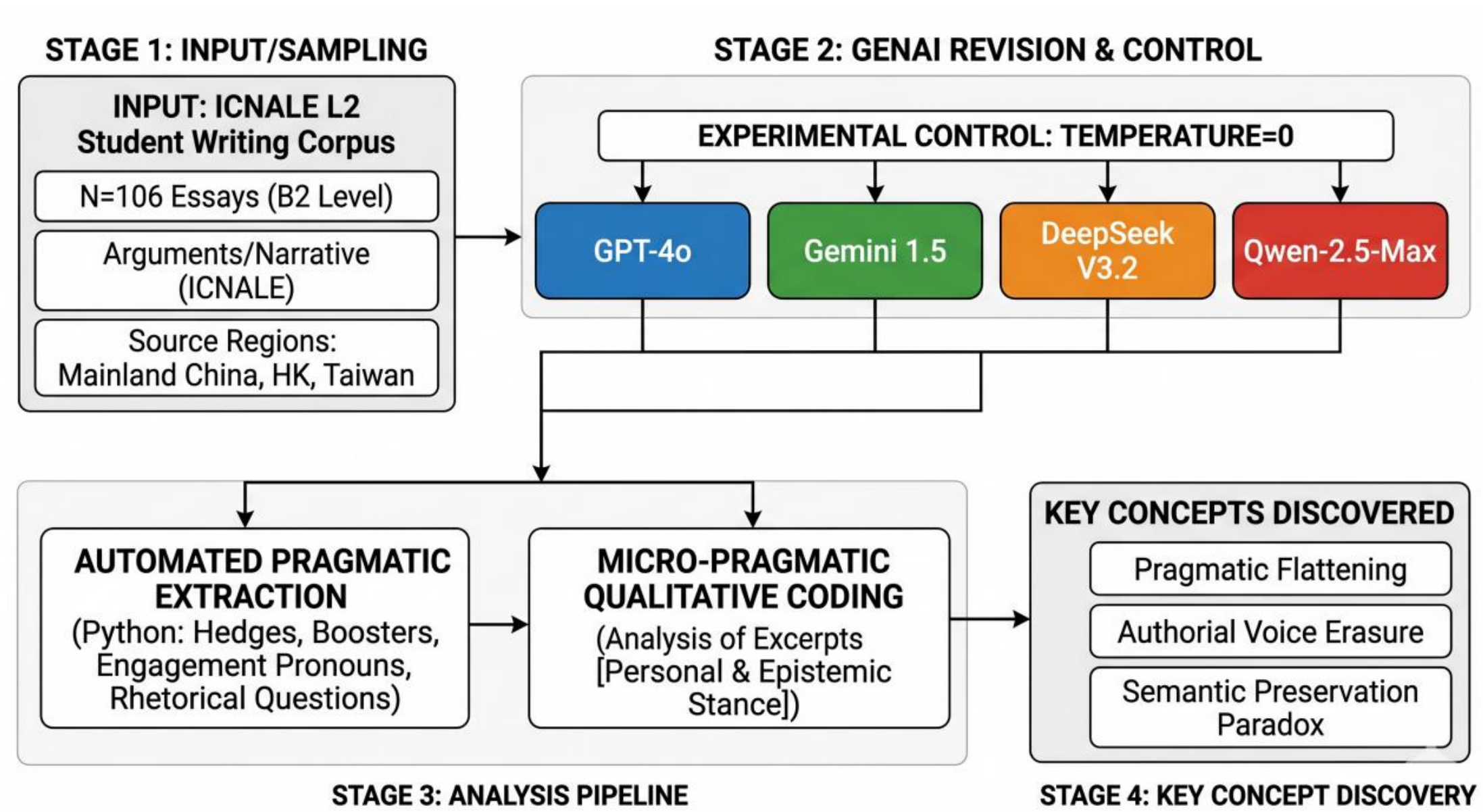

*Note:* The figure illustrates the integrated mixed-methods approach, including corpus sampling, GenAI revision control, and the dual-stream analysis pipeline.

### 3.1 Data Collection and Corpus Selection

To ensure the analysis is grounded in authentic and representative L2 discourse, a specialized dataset was extracted from the International Corpus Network of Asian Learners of English (ICNALE) (Ishikawa, 2013). The dataset comprises 106 argumentative essays authored by Chinese university students from three distinct regions: Mainland China, Hong Kong, and Taiwan. All selected writers are certified at the B2 proficiency level according to the Common European Framework of Reference for Languages (CEFR).

The B2 proficiency tier represents a critical developmental stage where L2 writers possess sufficient lexicogrammatical competence to articulate coherent arguments but may still rely on specific, culturally informed sociopragmatic strategies to construct their authorial voice. To maintain topic consistency, the 106 essays are evenly distributed across two standardized argumentative prompts within ICNALE: whether college students should have part-time jobs (PTJ) and whether smoking should be completely banned in all public places (SMK). These original, unmodified essays constitute the baseline "Human-Authored Group" (Group A).

### 3.2 Generative AI Interventions and Experimental Control

To assess whether pragmatic flattening is an algorithmic universal or model-specific bias, this study expanded the GenAI intervention to include four leading Large Language Models (LLMs): GPT-4o, Gemini 1.5, DeepSeek V3.2, and Qwen-2.5-Max. The inclusion of both Western-developed (GPT-4o, Gemini) and Eastern-developed (DeepSeek, Qwen) models allows for a comprehensive cross-algorithmic comparison.

The 106 original essays from Group A were systematically processed through the official APIs of these four models, resulting in 424 AI-generated parallel texts (a total corpus of 530 texts). Establishing rigorous experimental control over the LLMs is paramount for maintaining the validity of linguistic comparisons. Drawing on established methodologies for evaluating GenAI (Mizumoto & Eguchi, 2023), the temperature parameter for all four models was strictly set to 0. This zero-temperature setting minimizes stochastic randomness and hallucinations, ensuring deterministic, highly reproducible outputs based purely on the models' core normative algorithms.

To simulate authentic, real-world L2 user behavior, the following system prompt was applied uniformly across all 424 interactions:

*"Act as a professional academic English editor. Please comprehensively polish the following essay written by a non-native speaker. Eliminate unnatural phrasing, enhance flow, and ensure the text meets the standards of elite American/British academic discourse, but do not add new arguments."*

### 3.3 Computational Data Extraction and Quantitative Analysis

Due to the large volume of the parallel corpus (530 texts), a custom Python script utilizing Natural Language Processing (NLP) libraries was developed to automate the extraction and statistical computation of targeted pragmatic features.

The quantitative analysis was anchored in Hyland's (2005) framework of interactional metadiscourse. The Python script systematically parsed the corpus to calculate the raw frequencies of four distinct sociopragmatic variables:

Hedges: Markers of tentative epistemic stance used to mitigate claims (e.g., may, perhaps, it seems).

Boosters: Markers of authoritative assertion and certainty (e.g., must, undoubtedly, in fact).

Engagement Pronouns: Interactive markers used to build reader solidarity (e.g., we, you, our).

Rhetorical Questions: High-context dialogic strategies used to negotiate meaning.

To ensure strict comparability across essays of varying lengths, the Python script automatically normalized all raw frequencies to occurrences per 1,000 words. Furthermore, the script calculated the Mean and Standard Deviation (SD) for each variable across the human baseline and the four AI models. The variance calculation (SD) is specifically designed to quantitatively test the hypothesis of "cultural ghosting"—i.e., whether the models homogenize the regional rhetorical diversity of students from Mainland China, Hong Kong, and Taiwan into a singular pragmatic norm.

3.4 Micropragmatic Qualitative Coding Procedure

To complement the macro-level statistical trends and understand the contextual mechanics of algorithmic modifications, a targeted qualitative coding procedure, guided by Saldaña (2021), was applied to representative text pairs.

This qualitative phase was grounded in Politeness Theory (Brown & Levinson, 1987) to evaluate the modification of Face-Threatening Acts (FTAs). Researchers manually tracked how the algorithms altered specific protective strategies designed to save the reader's negative face. By aligning the computational corpus extraction with manual micro-comparisons of specific sentences (e.g., the algorithmic conversion of a tentative student suggestion into a hyper-direct machine directive), this mixed-methods framework provides a robust empirical foundation to fully articulate the Semantic Preservation Paradox.

# 4. Results

### 4.1 Macro-Quantitative Overview of Pragmatic Flattening

To empirically validate the structural shifts in sociopragmatic choices, a macro-quantitative analysis was conducted on the 530 parallel texts. The raw frequencies of four interactional metadiscourse markers were extracted and normalized to occurrences per 1,000 words. Table 1 summarizes the descriptive statistics (Means and Standard Deviations) across the human-authored baseline and the four GenAI models.

**Table 1. Descriptive Statistics of Pragmatic Markers Across Sources (Normalized per 1,000 words)**

| Source Model | Hedges M(SD) | Boosters M(SD) | Engagement M(SD) | Questions M(SD) |
|---|---|---|---|---|
| Human (Baseline) | 7.97 (8.62) | 9.26 (7.26) | 17.16 (22.86) | 1.11 (2.44) |
| DeepSeek v3.2 | 8.68 (7.09) | 8.59 (7.08) | 2.72 (6.61) | 0.83 (2.24) |
| GPT-4o | 7.88 (5.83) | 7.21 (5.12) | 1.92 (5.58) | 0.63 (1.56) |
| Gemini 1.5 | 3.98 (3.75) | 4.91 (4.31) | 4.55 (3.16) | 0.17 (0.65) |
| Qwen-2.5-Max | 6.50 (6.35) | 7.36 (6.80) | 3.77 (8.10) | 0.72 (1.96) |

The computational extraction reveals a dimensional divergence in the algorithmic processing of L2 sociopragmatic markers. The algorithms demonstrated bifurcated behaviors depending on the pragmatic function of the markers. The data shows a universal collapse in interactive engagement alongside model-specific variability in epistemic

**Figure 2: Mean Frequencies of Interactional Metadiscourse Markers Across Human and AI Conditions.stance, and is illustrated in Figure 2.**

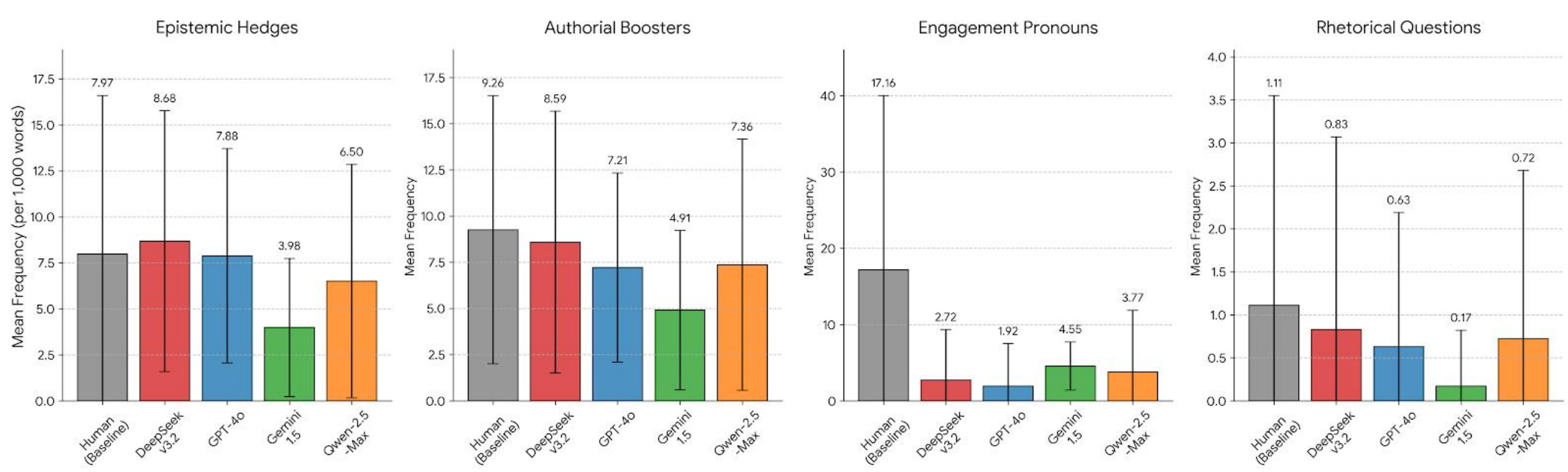

*Note:* This figure presents the normalized mean frequencies of four interactional metadiscourse marker categories across five distinct conditions, based on the descriptive statistics provided in Table 1. Error bars represent standard deviations, indicating stylistic variance within each group.

The interactive dimension exhibits a drastic and universal algorithmic erasure of dialogic engagement strategies. In the original human dataset, engagement pronouns appeared at a notably high frequency (Mean = 17.16). These pronouns function as vital devices for building conversational solidarity and negotiating meaning with the reader. Following GenAI intervention, this frequency collapsed across all models. GPT-4o exhibited the most aggressive reduction, plummeting to 1.92. The other models maintained similarly marginalized levels (DeepSeek = 2.72, Qwen = 3.77, Gemini = 4.55). The use of rhetorical questions dropped consistently and nearly vanished in Gemini's outputs (Mean = 0.17). LLMs systematically dismantle the negotiated, high-context conversational stance of L2 writers. Interactive discourse is converted into monologic assertions across all underlying algorithms.

In dimension of epistemic stance, the model exhibits highly varied, architecture-specific behaviors. This challenges the assumption of AI uniformly replacing tentative language with absolute certainty. As evidenced in Table 1, Gemini 1.5 exerted an intense dual-scrubbing effect. It roughly halved both tentative hedges (Mean = 3.98) and authoritative boosters (Mean = 4.91) compared to the human baseline (Hedges = 7.97, Boosters = 9.26). Other models reacted differently. GPT-4o maintained a near-baseline level of hedges (Mean = 7.88). DeepSeek v3.2 increased the raw frequency of hedges (Mean = 8.68) while slightly reducing boosters (Mean = 8.59).

This divergence substantiates a refined understanding of the Semantic Preservation Paradox. The algorithms universally strip away the protective face-saving mitigators of direct human interaction (pronouns and questions). They do not uniformly erase epistemic markers. They apply a decontextualized algorithmic normativity. This normativity manifests as the hyper-objective scrubbing seen in Gemini or the automated algorithmic caution injected by DeepSeek.

Furthermore, the standard deviation (SD) metrics provide mathematical evidence for the cultural ghosting hypothesis. The original human essays displayed significant stylistic variance (e.g., Engagement SD = 22.86), reflecting diverse rhetorical preferences. Post-intervention, the SDs for all pragmatic markers were drastically compressed across all four LLMs. This concurrent compression indicates GenAI acts as a homogenizing agent. The natural linguistic diversity of L2 writers is flattened into a rigid, identical machine-generated voice.

**4.2 Micropragmatic Qualitative Analysis: The Erasure of Authorial Voice**

While the quantitative data establishes the macroscopic trend of pragmatic flattening, a micro-level qualitative comparison of parallel texts reveals the specific algorithmic mechanics behind the "Semantic Preservation Paradox." By examining how GenAI processes highly subjective, high-context L2 writing, we can observe the systematic overwriting of student identity.

Excerpt 1 (Table 2) illustrates a typical introduction from a Chinese L2 writer discussing a smoking ban, juxtaposed with its AI-generated revisions.

**Table 2. Qualitative Comparison of Dialogic Engagement and Authorial Stance**

| Source | Excerpt from Introduction / Stance Formulation |
|---|---|
| Human (Original) | "*Should smoking be banned at all the restaurants*? About this, different people hold different views. ... *In my opinion, I agree with it... I hate them* who smoke at the restaurant. Firstly it does harm to everyone. Secondly, *I think* it is an irresponsible behavior." |
| GPT-4o | "*The question of whether smoking should be banned in all restaurants has sparked considerable debate*. ... Smoking in dining establishments exposes non-smokers to secondhand smoke, which is universally acknowledged as a severe health hazard." |
| Qwen-2.5-Max | "*... the ban on smoking in all restaurants is a necessary measure* to protect public health, ensure a pleasant dining experience, and promote a cleaner environment." |

*Note: Italics added to highlight pragmatic markers and epistemic stance.*

In the original text, the L2 writer relies heavily on dialogic engagement and overt subjective stance to construct their argument. The essay opens with a direct rhetorical question (*"Should smoking be banned...?"*), a strategy frequently employed in high-context cultures to invite reader participation and negotiate common ground before delivering a thesis. Furthermore, the writer's epistemic stance is highly visible and emotionally charged, utilizing explicit first-person boosters and mental verbs (*"In my opinion, I agree," "I hate," "I think"*).

The GenAI interventions, however, demonstrate a profound intolerance for this conversational and subjective rhetorical style. As seen in the GPT-4o revision, the direct rhetorical question is algorithmically flattened into a declarative, third-person noun phrase (*"The question of whether... has sparked considerable debate"*). The emotionally resonant, first-person subjective claims (*"I hate," "I think"*) are entirely scrubbed and replaced with hyper-objective, institutionalized academic phrasing (*"universally acknowledged as a severe health hazard"*).

This transformation perfectly exemplifies the Semantic Preservation Paradox: the core propositional meaning (that smoking is harmful and should be banned) is retained and even lexicogrammatically enhanced, yet the L2 writer's unique sociopragmatic identity—their passion, their dialogic invitation to the reader, and their overt personal stance—is completely erased. The AI essentially acts as an agent of linguistic standardization, enforcing a singular, depersonalized Anglo-American academic norm regardless of the original author's rhetorical intent.

While Excerpt 1 highlights the algorithmic erasure of dialogic questions, Excerpt 2 (Table 3) illustrates a complementary facet of pragmatic flattening: the rigidification of tentative, idiosyncratic L2 stance into homogenized, institutionalized directives. Here, a Hong Kong L2 writer discusses the viability of university students taking part-time jobs.

**Table 3. Qualitative Comparison of Epistemic Stance and Directive Homogenization**

| Source | Excerpt on Part-time Jobs / Stance Formulation |
|---|---|
| Human (Original) | "...*I am for* students to pick up part time jobs to cover for their expenses. ... *I feel it is not advisable* for a student to pick up part time work as the time available should be fully utilize..." |
| Qwen-Max (AI Polish) | "...*I strongly advocate* for those who need to work to cover their expenses. ... Beyond these two scenarios, *I do not generally advise* students to take on part-time work." |

*Note: Italics added to highlight pragmatic markers and epistemic stance.*

In the baseline text, the student's authorial voice is negotiated and subtly hedged. Phrases such as *"I am for"* and *"I feel it is not advisable"* reflect a highly personalized, slightly tentative epistemic stance. These choices protect the reader's negative face by framing the argument as a personal feeling rather than an absolute, universal mandate.

However, the AI intervention fundamentally alters the pragmatic force of these statements. As evidenced in the Qwen-Max output, the student's personal feelings are structurally converted into authoritative, high-certainty directives. The somewhat colloquial but authentic *"I am for"* is escalated into the forceful institutional booster *"I strongly advocate."* Similarly, the hedged expression of opinion *"I feel it is not advisable"* is hardened into a formal, quasi-expert prohibition: *"I do not generally advise."*

This qualitative shift provides crucial micro-level validation for the phenomenon of "cultural ghosting." Even when the AI refrains from completely deleting the first-person pronoun, it overwrites the L2 writer’s authentic, negotiated cadence with an inflexible, formulaic Anglo-American academic persona. The AI essentially treats the student's natural sociopragmatic variation as an "error" to be corrected, thereby standardizing the text at the cost of the author's unique rhetorical identity.

## 5. Discussion

The mixed-methods findings of this study confirm that the integration of Generative AI in L2 writing extends far beyond mere lexicogrammatical error correction. By systematically intervening in the sociopragmatic layer of text, LLMs trigger profound shifts in how multilingual writers negotiate meaning and project their authorial voice.

### 5.1 Decoding the Semantic Preservation Paradox and Neutrality Bias

Evidence from the parallel corpus analysis substantiates the presence of a Semantic Preservation Paradox across all four generative models. While the Large Language Models (LLMs) demonstrated a high degree of efficacy in retaining the core propositional meaning of the L2 argumentative essays, they simultaneously engaged in a systematic eradication of interpersonal metadiscourse. Quantitative findings indicate a universal and drastic collapse in the frequency of engagement pronouns and rhetorical questions. Such a trend suggests that current GenAI algorithms prioritize a monologic mode of delivery over the interactive negotiation of meaning that is characteristic of human student writing.

A significant finding of this investigation centers on the Dimensional Divergence within the sociopragmatic layer. Contrary to the assumption that AI polishing uniformly replaces tentative language with absolute certainty, the models exhibited multifaceted behaviors regarding epistemic stance. Specifically, Gemini 1.5 and GPT-4o demonstrated a pronounced neutrality bias by scrubbing both face-saving mitigators and subjective authorial claims. In the case of Gemini 1.5, the frequency of hedges was reduced to 3.98 while boosters fell to 4.91, indicating a severe dual-scrubbing effect. This dual process transforms interactive student writing into depersonalized, hyper-objective statements.

Interestingly, data from DeepSeek v3.2 reveals a distinct algorithmic trajectory where the frequency of hedges increased to 8.68 compared to the human baseline of 7.97. This deviation from the general scrubbing trend implies that certain architectures may be programmed to reinforce a form of decontextualized algorithmic caution. This supplementary hedging likely serves as a defensive linguistic mechanism intended to satisfy internal alignment parameters for objectivity. Consequently, regardless of whether the AI removes or adds hedges, the result is a profound loss of interpersonal agency. The models effectively treat overt authorial voice as stylistic noise to be filtered out in favor of an institutionalized academic norm.

### 5.2 Algorithmic Hegemony and the Evidence of Cultural Ghosting

Robust mathematical evidence for the Cultural Ghosting hypothesis is found in the drastic compression of standard deviation (SD) metrics across all pragmatic variables.

The original human-authored baseline displayed significant stylistic variance, particularly in the use of engagement markers with a standard deviation of 22.86. This variance reflects the diverse rhetorical and cultural backgrounds of students from Mainland China, Hong Kong, and Taiwan. Following the AI interventions, this natural sociopragmatic diversity was mathematically flattened across all four models. For instance, Gemini reduced the engagement variance to a mere 3.16, representing an aggressive homogenization of the L2 writer's original voice.

Such a concurrent compression of standard deviations indicates that GenAI functions as a powerful homogenizing agent. This process enforces a singular, rigid pragmatic norm that erases regional rhetorical fingerprints in favor of an identical machine-generated voice. Moreover, the convergence of Western-developed models like GPT-4o and Eastern-developed models like DeepSeek and Qwen on an identical low-engagement paradigm exposes the depth of digital hegemony. All four models transitioned the student texts toward a monologic, depersonalized rhetorical paradigm when tasked with academic polishing.

This cross-model consistency suggests that the underlying training data and alignment processes are rooted in the same normative conventions. The algorithms interpret elite academic discourse as inherently devoid of high-context interactional markers. By marginalizing the rhetorical strategies of learners from high-context cultures, the models act as agents of linguistic standardization. Multilingual learners from the ICNALE corpus are effectively stripped of their unique rhetorical identities, leading to a structural enforcement of Anglo-American academic templates.

### 5.3 The Ecological Validity of the Prompt and Internalized Native-Speakerism

Interpreting the severity of this pragmatic flattening requires an analysis of the specific system prompt utilized in the study. The instruction to meet the standards of elite American or British academic discourse likely triggered a specific algorithmic response. This prompt design secures high ecological validity as it simulates the authentic behavior of L2 students who often request AI to make their writing sound native or professional. Such requests are frequently driven by an internalized linguistic insecurity or native-speakerism.

The AI's response to these prompts reveals a problematic equation of elite writing with the total eradication of subjective and interactive metadiscourse. When L2 writers delegate their textual agency to AI in pursuit of native-like perfection, they unwittingly consent to the erasure of their cultural identities. This dynamic highlights a profound ethical dilemma in the digital age. Algorithmic obedience to the restrictive framing of the prompt forces the AI to prioritize a depersonalized persona over the writer's original intent. Consequently, the uncritical adoption of GenAI for language polishing becomes a process of structural marginalization.

### 5.4 Pedagogical Implications Toward Critical AI Literacy

The systemic erasure of authorial voice necessitates a paradigm shift in how AI tools are integrated into language education. Relying on LLMs solely as automated proofreaders is pedagogically risky. This approach implicitly validates the algorithmic flattening of diverse rhetorical expressions. To counter this trend, writing curricula must prioritize the development of Critical AI Literacy. Educators should guide students to deconstruct algorithmic bias by comparing their original drafts with AI-generated outputs.

Furthermore, the curriculum should incorporate voice-preserving prompt engineering. Students must be taught to explicitly command the AI to maintain their original rhetorical questions and first-person perspectives while correcting grammatical errors. Finally, language programs need to actively rebuild students' rhetorical confidence. Culturally informed sociopragmatic strategies should be framed as valuable assets in global academic discourse rather than as errors to be corrected. Such interventions will empower multilingual writers to negotiate their identities within hybrid human-AI writing environments.

## 6. Conclusion

### 6.1 Summary of Main Findings

This study systematically investigated the impact of generative artificial intelligence on the sociopragmatic dimensions of L2 academic writing. By employing a mixed-methods design to analyze 530 parallel texts across a human baseline and four prominent Large Language Models (LLMs), the research provides robust empirical evidence for the phenomena of pragmatic flattening and cultural ghosting. The findings articulate a refined understanding of the Semantic Preservation Paradox. While GenAI algorithms excel at optimizing lexicogrammatical accuracy and retaining core propositional meaning, they achieve this by systematically eradicating the author's interpersonal metadiscourse.

The data reveals a critical Dimensional Divergence in algorithmic intervention. In the interactive dimension, all models demonstrated a universal and drastic collapse of dialogic engagement strategies. Specifically, engagement pronouns, which serve as vital devices for building conversational solidarity, plummeted from a human baseline mean of 17.16 to marginalized levels between 1.92 and 4.55. However, in the dimension of epistemic stance, the models exhibited architecture-specific variability. While models like Gemini exerted an aggressive dual-scrubbing effect on both hedges and boosters, DeepSeek v3.2 demonstrated a form of "algorithmic caution" by increasing the frequency of hedges to 8.68.

### 6.2 Theoretical and Ethical Implications

The systematic erasure of authorial voice by generative algorithms indicates a structural enforcement of digital hegemony. Driven by an underlying neutrality bias and an internalized native-speakerism, the models consistently dismantled the negotiated, high-context conversational stance of L2 writers. The standard deviation metrics provide robust mathematical evidence for the Cultural Ghosting hypothesis. The original human essays displayed significant stylistic variance, such as an Engagement SD of 22.86, reflecting diverse rhetorical preferences among students from different regions. Post-intervention, these SDs were drastically compressed across all LLMs, with Gemini reducing variance to 3.16.

This compression suggests that GenAI acts as a homogenizing agent, enforcing a singular, rigid pragmatic norm that flattens natural linguistic diversity into an identical machine-generated voice. This algorithmic intervention is not a culturally neutral process but a structural enforcement of Anglo-American academic templates that marginalizes the authentic voices of multilingual learners. When L2 writers delegate their textual agency to AI in pursuit of native-like perfection, they unwittingly consent to the erasure of their unique cultural and rhetorical identities.

### 6.3 Pedagogical Recommendations

The findings of this study necessitate a paradigm shift in the integration of AI tools within language education. Relying on LLMs solely as automated proofreaders is pedagogically risky, as it implicitly validates the algorithmic flattening of diverse rhetorical expressions. To counter this, writing curricula must prioritize the development of Critical AI Literacy. Educators should guide students to deconstruct algorithmic bias by actively comparing their original drafts with AI-generated outputs to visualize how their personal stance is systematically altered.

Furthermore, the curriculum must incorporate voice-preserving prompt engineering. Students should be explicitly taught to command the AI to fix grammatical errors while strictly maintaining their original rhetorical questions and first-person perspectives. Finally, language programs need to actively rebuild students' rhetorical confidence by emphasizing that culturally informed sociopragmatic strategies are valuable assets in global academic discourse rather than unnatural errors.

### 6.4 Limitations and Directions for Future Research

While this study offers critical insights, several limitations should be acknowledged. First, the corpus was restricted to argumentative essays authored by B2-level Chinese learners of English. The pragmatic flattening effect may manifest differently in other genres or among learners from varied linguistic backgrounds. Second, the experimental design utilized a directive prompt targeting "elite academic discourse," which introduces a potential confounding variable regarding algorithmic obedience.

Future research is encouraged to investigate how the Semantic Preservation Paradox operates across varying L2 proficiency levels, from A2 beginners to C1 advanced learners. Scholars should also incorporate "neutral prompt" control groups to isolate the specific locus of algorithmic homogenization. Additionally, longitudinal classroom-based studies are needed to evaluate how the explicit teaching of Critical AI Literacy impacts students' long-term rhetorical confidence and their ability to negotiate identity in hybrid human-AI writing environments.

## Declaration of Generative AI in Visual Content:

During the preparation of this work, the author used Google Gemini (specifically, the Gemini 3 Flash Image model) to assist in generating Figure 1 and Figure 2. After using this tool, the author reviewed and refined the generated images, and takes full responsibility for the final content of the publication.